\documentclass{article}
\usepackage{ijcai24}

\usepackage{times}
\usepackage{soul}
\usepackage{url}
\usepackage[hidelinks]{hyperref}
\usepackage[utf8]{inputenc}
\usepackage[small]{caption}
\usepackage{graphicx}
\usepackage{amsmath}
\usepackage{amsthm}
\usepackage{booktabs}
\usepackage{algorithm}
\usepackage{algorithmic}
\usepackage[switch]{lineno}
\usepackage{xcolor}

\title{Towards Automatic Composition of ASP Programs\\  from Natural Language Specifications\footnote{This work has been submitted to a conference and is awaiting approval}}

\author{
Manuel Borroto$^1$
\and
Irfan Kareem*$^1$\and
Francesco Ricca$^{1}$
\affiliations
$^1$Department of Mathematics and Computer Science, University of Calabria, Rende CS, 87036, Italy\\
\emails
\{manuel.borroto,irfan.kareem,francesco.ricca\}@unical.it
}

\begin{document}

\maketitle

\begin{abstract}

This paper moves the first step towards automating the composition of Answer Set Programming (ASP) specifications.
In particular, the following contributions are provided: 
(i) A dataset focused on graph-related problem specifications, designed to develop and assess tools for ASP automatic coding;
(ii) A two-step architecture, implemented in the NL2ASP tool, for generating ASP programs from natural language specifications.
NL2ASP uses neural machine translation to transform natural language into Controlled Natural Language (CNL) statements. Subsequently, CNL statements are converted into ASP code using the CNL2ASP tool.
An experiment confirms the viability of the approach.

\end{abstract}

\section{Introduction}
Natural Language Processing (NLP) methodologies~\cite{DBLP:books/lib/JurafskyM09} have determined impressive changes in the way we engage with computers, streamlining a multitude of tasks that would otherwise demand considerable time and proficiency from users. 
A noteworthy example is the emergence of tools for the automatic composition of computer programs~\cite{DBLP:journals/software/ErnstBM22,DBLP:journals/corr/abs-2302-06590}, like GitHub Copilot~\cite{DBLP:journals/jss/DakhelMNKDJ23,github}.
Nevertheless, prevailing tools of this nature predominantly cater to widely used programming languages, and limited (or no support) is available for many problem-solving formalisms in the area of Knowledge Representation and Reasoning (KR\&R).
An example of formalism for KR\&R that misses automatic program composition tools is Answer Set Programming (ASP).

Answer Set Programming (ASP)~\cite{DBLP:journals/cacm/BrewkaET11,DBLP:conf/iclp/GelfondL88} is a declarative programming paradigm that can be used to solve complex AI problems. 
ASP originates from research on logic programming, nonmonotonic reasoning and knowledge representation. 
ASP became popular for featuring both an expressive declarative language and some efficient implementations~\cite{DBLP:journals/tplp/GebserMR20}, such as Clingo~\cite{DBLP:conf/iclp/GebserKKOSW16} and DLV~\cite{DBLP:conf/lpnmr/AlvianoCDFLPRVZ17}. 
ASP has been applied both in academia and industry, and it proved to be effective in several knowledge-intensive applications of AI, such as scheduling, product configuration, robotics, workforce management, and decision support~\cite{DBLP:journals/aim/ErdemGL16,DBLP:journals/tplp/GebserMR20}. 

In the last few years, quite some effort has been spent on providing programming environments and tools for assisting programmers in devising ASP specifications~\cite{DBLP:journals/tplp/VosKOPT12,DBLP:conf/iclp/AlvianoCR23,DBLP:journals/corr/abs-2303-10118,DBLP:conf/lpnmr/FebbraroRR11,DBLP:journals/tplp/BusoniuOPST13}, including advanced editors, debuggers, testing frameworks, visualization tools, etc. 
Nonetheless, coding in ASP still results in very challenging for beginners, especially for those having a weak mathematical and logical background, and is a time-consuming (sometimes repetitive) task also for expert programmers.
On the other hand, once we look for the latest tools that significantly boosted the productivity of developers using mainstream imperative programming languages, it is easy to observe that a great benefit has been brought by the introduction of automatic program composition tools, such as GitHub Copilot~\cite{DBLP:journals/jss/DakhelMNKDJ23,github}.
However, no programming environment for ASP (as far as we know) features automatic code composition from the user's requests in natural language. 
One might observe that generative AI tools, like ChatGPT, are capable of generating some ASP code snippets from specific prompts. Actually, they have been proficiently used to produce factual statements in an ASP-based system for reasoning from text~\cite{DBLP:conf/acl/YangI023}; but, it can be easily verified that these AI models are not robust in generating valid and correct ASP programs. Thus, the problem of (robustly) composing ASP programs from textual statements provided in natural language remains open.

In this paper, the first steps towards filling this gap are taken, and the following contributions are provided: 

\begin{itemize}
    \item [(i)] A dataset focused on graph-related problem specifications, that is conceived to develop and assess NLP tools for the automatic composition of ASP programs;
    \item [(ii)] A two-step architecture, implemented in the NL2ASP tool, aiming at automating the translation of natural language specifications into ASP programs;
    \item [(iii)] An experiment providing empirical evidence that the dataset (of point $(i)$) and the architecture of point $(ii)$) are effective.
\end{itemize}

NL2ASP processes the input in two distinct phases: The first phase draws inspiration from Neural Machine Translation (NMT). 
NMT~\cite{DBLP:journals/corr/abs-2002-07526} methods utilize deep learning algorithms to translate text from one language to another, providing more natural and accurate results compared to traditional machine translation methods \cite{DBLP:journals/jair/Stahlberg20}.
NL2ASP employs NMT techniques to translate the input sentences in natural language to statements conforming to a recently-introduced controlled natural language for ASP~\cite{DBLP:conf/datalog/DodaroMR22}. 
For this task, NL2ASP can resort either to BART-base~\cite{DBLP:journals/corr/abs-1910-13461} or T5-small~\cite{DBLP:journals/jmlr/RaffelSRLNMZLL20}.
In the second phase, the Controlled Natural Language (CNL) statements are converted into ASP code using the CNL2ASP tool~\cite{DBLP:conf/datalog/DodaroMR22}. 
The performance of NL2ASP is measured in an experiment in terms of the quality of the statements produced in each of the steps of the composition process.
In particular, the CNL statements produced by NL2ASP are evaluated in terms of BLEU~\cite{DBLP:conf/acl/PapineniRWZ02}, METEOR~\cite{DBLP:conf/acl/BanerjeeL05} and Syntax Correctness Accuracy (a measure of the capability of producing syntactically-valid CNL statements). 
Eventually, the overall behavior of the system is assessed by implementing an end-to-end test in which the capability of producing correct ASP specifications is measured.
The obtained results are very promising and confirm the viability of the approach. 

\section{Related Work} 
The benefits and drawbacks of automated composition of programs have been studied in the literature~\cite{DBLP:journals/software/ErnstBM22}, and the enormous potential of these tools is nowadays recognized~\cite{github,DBLP:journals/corr/abs-2302-06590,DBLP:journals/jss/DakhelMNKDJ23}. Almost all mainstream programming languages have been supported by academic or industrial tools~\cite{DBLP:journals/corr/abs-2107-03374}. 
The benefits of providing tools for easing the development of ASP programs reducing the impedance mismatch existing from natural language specifications and ASP source code has been also recognized in the literature~\cite{DBLP:conf/bionlp/ErdemY09,DBLP:conf/inap/FangT17,DBLP:journals/tplp/Schwitter18,DBLP:conf/datalog/DodaroMR22}. 
We are not currently aware of ASP-specific tools targeting automatic program composition from natural language based on NMT as NL2ASP. Nonetheless, there are some that support inputs from more structured or simpler languages.

Baral et al. automated the solving of logic puzzles in simplified English by translating their descriptions into ASP~\cite{DBLP:conf/aaaifs/BaralD11} by resorting to $\lambda$-calculus and probabilistic combinatorial categorical grammars. 

The Controlled Natural Languages (CNL)s are subsets of the full natural languages with restricted grammar and vocabulary \cite{DBLP:journals/coling/Kuhn14}. 
Several efforts have been made to develop CNLs that target ASP programs. 
Erdem et al. described the specific grammatical structure of a CNL named BIOQUERYCNL, as well as the algorithm for converting queries into ASP~\cite{DBLP:conf/bionlp/ErdemY09}. 
Fang et al. introduced a CNL approach for representing answer sets based on LANA annotations~\cite{DBLP:conf/inap/FangT17} which was implemented in the SeaLion IDE. 
Schwitter developed in 2018 a grammar to specify and verbalize answer set programs using a CLN named PENG\textsuperscript{ASP}~\cite{DBLP:journals/tplp/Schwitter18}.
Lifschitz highlighted the connection between mathematical definitions and the knowledge representation capability of ASP~\cite{DBLP:conf/iclp/Lifschitz22}. 
More recently, Dodaro et al. presented CNL2ASP, a comprehensive publicly-available tool that converts controlled natural language into ASP programs \cite{DBLP:conf/datalog/DodaroMR22}. 
NL2ASP resorts to the CNL and tool devised by Dodaro et al.
CNLs ease the task of programming but still require human developers. NL2ASP aims at overcoming this by making the encoding process automatic.

Our approach is also related to the development of datasets for knowledge-based question answering and to other applications of ASP to natural language.
Datasets that are useful for question answering were proposed by 
Perevalov et al. that extended the Knowledge Graph Question Answering (KGQA) benchmarks QALD-9 by adding question query pairs up to 8 languages~\cite{DBLP:conf/semco/PerevalovDUB22}. Dubey et al. introduced a comprehensive dataset for complex question answering, called LC-QuAD 2.0 \cite{DBLP:conf/semweb/DubeyBA019}. This dataset includes 30,000 questions, their corresponding paraphrases, and SPARQL queries. 

A NMT method based on long short-term memory (LSTM) units was presented by Sutskever et al. on an English-to-French translation task \cite{DBLP:conf/nips/SutskeverVL14}. 
As part of the English-to-French translation task, Bahdanau et al. proposed a NMT module based on Bidirectional Recurrent Neural Networks \cite{DBLP:journals/corr/BahdanauCB14}. 
A NMT method for KBQA that automatically translates natural language in SPARQL queries was proposed by Borroto et al.~\cite{DBLP:journals/eswa/BorrotoR23} that performed well in the QUALD competition.

More recently, Nye et al. presented GPT-3 based dual-system model, which generates the semantic parses from natural language sentence and couples it with reasoning modules \cite{DBLP:conf/nips/NyeTTL21}. In this line of research, Yang et al. proposed LLMs like GPT-3 can act as few-shot semantic parsers, converting natural language into logical forms for answer set programs (ASP) \cite{DBLP:conf/acl/YangI023}. 
This system can address diverse question-answering tasks without distinct retraining.
Ishay et al. leveraged LLMs to get the ASP with prompt engineering for solving logic puzzles \cite{DBLP:conf/kr/IshayY023}. They have used the logic puzzle dataset of \cite{DBLP:conf/aaai/MitraB16}. Although LLMs excel in System-1 processes, their results can frequently be inconsistent and incoherent. 
%
%
%
Mitra et al. proposed a heterogeneous agent model for question-answering tasks in Facebook’s bAbl dataset \cite{DBLP:conf/aaai/MitraB16} where ASP is used for knowledge representation and reasoning. Moldovan et al. demonstrated that a logic prover can be integrated into a Question Answering system \cite{DBLP:conf/naacl/MoldovanCHM03}, where logic representations are used to transform questions and answers, axioms are provided to a prover. 

\section{Basics on Answer Set Programming}
Answer Set Programming (ASP)~\cite{DBLP:journals/cacm/BrewkaET11,DBLP:journals/ngc/GelfondL91} is a declarative logic-based language for knowledge representation and reasoning. 
Rather than writing algorithms, the ASP programmer describes problems with a declarative formal language and uses an ASP system~\cite{DBLP:journals/aim/LierlerMR16} to find solutions~\cite{DBLP:books/sp/Lifschitz19}.

\paragraph{The language of ASP.}
The main construct of the ASP language is the logic rule, which is a construct of the form:
\begin{equation}
a_{1} \ | \ldots | \ a_{n} \ :- \; b_{1}, \ldots, b_{k} \; ; \; not \; b_{k+1}, \ldots, \; not \; b_{m}
\end{equation}

\noindent where $a_i$ ($0 \leq i \leq n$) are \textit{atoms}, $b_i$ ($0 \leq i \leq k \leq m$) and $not b_j$ ($k \leq j \leq m$) are positive and  negative \textit{literals}, respectively. 
Informally, a rule reads as follows: ``at least one of the $a_i$ is true whenever all $b_{i}$ ($0 \leq i \leq k \leq m$) are true and all  $b_j$ ($k \leq j \leq m$) are false''.
On the right-hand side of a rule is the \textit{body} (or antecedent) and on the the left-hand side of a rule is the \textit{head} (or consequent). The rule is called \textit{Fact} if the body of a rule is empty. A rule with an empty head is called \textit{Constraint}. 
An ASP program is a set of logic rules, which is interpreted according to common sense principles. An ASP program represents a problem to be solved that together with input (also expressed by a collection of rules) admits a collection of answers (possibly also no answer) corresponding to the solutions of the problem~\cite{DBLP:books/sp/Lifschitz19}.

For example, the well-known 3-colorability is solved by:
$$col(X,red) \ | \ col(X,green)\  | \ col(X,yellow) \ :- \; node(X)$$
$$:- col(X,C), col(Y,C), edge(X,Y) $$

The first rule reads ``if X is a node then it can be colored either in red or in green or in yellow'', and the second rule (a constraint) reads ``it is not possible that adjacent nodes  have the same color''. Here, the input is modelled by a set of facts of the form $node(\cdot)$ and $edge(\cdot,\cdot)$.
Suppose we provide in input the graph $node(1)$, $node(2)$, $node(3)$, $edge(1,2)$ and $edge(1,3)$; it can be verified that a possible solution (answer set) is $\{col(1,red), col(2,green), col(3,green)\}$.





ASP also supports weak constraints, choice rules, aggregates, etc.; however a full description of the ASP language is not instrumental for this paper. Please refer to dedicated literature~\cite{DBLP:journals/tplp/CalimeriFGIKKLM20,DBLP:journals/cacm/BrewkaET11,DBLP:journals/ngc/GelfondL91,DBLP:books/sp/Lifschitz19,DBLP:series/synthesis/2012Gebser} for a more detailed (and formal) account on ASP.


\paragraph{Controlled natural language for ASP.}
Controlled Natural Languages (CNL)s are subsets of natural languages with restricted grammar and vocabulary \cite{DBLP:journals/coling/Kuhn14}. 

A comprehensive CNL tool for ASP that supports all the main language constructs is CNL2ASP~\citeauthor{DBLP:conf/datalog/DodaroMR22}.
%
%
%
In CNL2ASP, CNL specifications are made up of propositions that are structured by means of clauses, connected by connectives to express concepts and conditions. 
To provide flavour and a basic understanding of CNL2ASP we resort to our running example, i.e., and encoding of the 3-colorability problem. 
First of all, we define node, edge and color by using the following definition statements: 

\texttt{\small A node is identified by an id.}

\texttt{\small A edge is identified by a firstnode, and by a secondnode.}

\texttt{\small A color is identified by an id.}

These are used by CNL2ASP to initialize the internal data structures and know about the domain of discourse.
Then, we specify the assignment of colors to nodes by using the following \texttt{whenever/then} clause:

\texttt{\small Whenever there is a node with id X then we can have a col with node X, and with color equal to blue, or a col with node X, and with color equal to red, or a col with node X, and with color equal to green.}

Finally, we specify valid colorings with the following \textit{negative strong constraint} statement:

\texttt{\small It is prohibited that C1 is equal to C2, whenever there is a col with node X, and with color C1, whenever there is a col with node Y, and with color C2, whenever there is an edge with firstnode X, and with secondnode Y.}

The ASP program resulting from a call to CNL2ASP is:
$$col(X,blue)\ |\ col(X,red)\ |\ col(X,green)\ :-\ node(X).$$
$$:- C1 = C2,\ col(X,C1),\ col(Y,C2),\ edge(X,Y).$$

It is worth observing that
the CNL specification is amenable to a human and does not require to know the technicalities of the syntax of ASP.
The interested reader can find more details CNL2ASP in the original article~\cite{DBLP:conf/datalog/DodaroMR22} and in the Github repository of CNL2ASP.%
\footnote{\url{https://github.com/dodaro/cnl2asp}} 

\section{The problem and some assumptions}

The problem we tackle in this paper is to ease the process of composing formal ASP statements (ASP programs) from problem specifications given in natural language.
However, this is a quite ambitious final objective. 
In order to take the first steps towards pursuing this goal we begin by making some pragmatic simplifying assumptions.

The first assumption we make is: problem formulations are given as a bag of statements, where contiguous subsets of statements can be encoded with a logical rule.
This makes it possible to compose a program by transforming iteratively group of statements in the corresponding rule. 

The second assumption we make is that we are given a corpus of pairs that associates each bag of statements to a \textit{``gold program"}. 
The gold program is an example of an expected encoding of the statements in ASP.
A subset of that corpus can be used to \textit{train} a neural network to perform the task of ASP program composition (training set); another (disjoint from training set) subset of that corpus can be used for testing correctness of the system (test set).

The problem we tackle is, thus,  formalized as follows:
Given a specification $P$ of a problem (a bag of statements), compose a program $P_{\text{asp}}$ that is uniform equivalent to the gold program $P_{\text{gold}}$ associated to $P$.

Two ASP programs $P_1$ and $P_2$ are uniform equivalent iff for any set of (non-disjunctive) facts $F$, $P_1 \cup F$ and $P_2 \cup F$ admit the same answer sets~\cite{DBLP:journals/tocl/EiterFW07}.
This captures that an ASP program models uniformly a problem over varying instances provided via facts~\cite{DBLP:journals/cacm/BrewkaET11}.

The formulation of the composition problem makes it possible to process iteratively the input by identifying the statements corresponding to a rule, and transforming such statements into the corresponding rule. 
In our approach, this latter task is performed in two steps: $(i)$ translation of the natural language in a corresponding CNL (NMT task); $(ii)$ conversion of the CNL in ASP (basic task of CNL2ASP).

To the best of our knowledge, no dataset supporting the above tasks is available; thus, we have developed one as described in the next section.
Pragmatically, we concentrated our efforts on graph-related problems. This is a domain of problems that is classically addressed with ASP~\cite{DBLP:journals/cacm/BrewkaET11,DBLP:journals/aim/ErdemGL16,DBLP:journals/aim/LierlerMR16,DBLP:journals/tplp/CalimeriFGIKKLM20}, and is very rich and variegate (in terms of number, complexity and importance of problems) to be considered a meaningful pragmatic choice to assess our approach.

Concerning the issue of testing our implementation, we observe that checking uniform equivalence of ASP programs is intractable (e.g., it belongs to the second level of the polynomial hierarchy for ground disjunctive ASP programs)~\cite{DBLP:journals/tocl/EiterFW07}.
Thus, in our experiment we will pragmatically consider, in analogy to what has been done in other similar contexts, looser measures of correctness, such as purely syntactic measures (BLEU, METEOR); and, in end-to-end tests, correctness was certified by a human expert.

\section{Dataset Construction}
\begin{figure}[t]
\centering
\includegraphics[width=0.9\columnwidth,trim={ 0 12 0 3 },clip]{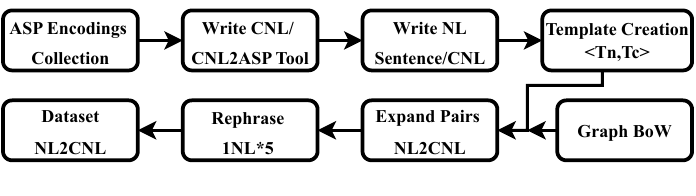} 
\caption{Dataset Creation workflow}.
\label{fig:datasetworkflow}
\end{figure}

One of the contributions of our work is a specialized dataset centered on graph-related problem specifications. This section details the creation process of this dataset, named NL2CNL. 
The workflow of dataset creation is illustrated in Figure~\ref{fig:datasetworkflow}, which entails collecting ASP encodings of graph-related problems.
NL2CNL includes problem statements in natural language and corresponding CNL propositions.
%

The ASP encodings were mainly collected from ASP competitions (2015, 2017), books, lecture notes, data shared by renowned professors, and other online sources.
%
%
For each collected ASP encoding, we have written the corresponding CNL and generated a new ASP encoding by using the CNL2ASP~\cite{DBLP:conf/datalog/DodaroMR22} tool. Then, we cross-verified the answer sets of both actual ASP encoding and tool-generated ASP encoding to identify trivial problems, and checked the encoding manually. This way, we ensure the CNL we have written is compliant with the actual encoding.
We have tried to cover the five main grammar propositions of tool CNL2ASP: (i) negative strong constraint proposition, (ii) positive strong constraint proposition, (iii) weak constraint proposition, (iv) definition proposition, (v) quantified choice proposition. We have ensured that each CNL is accompanied by a corresponding natural language statement that accurately reflects the original semantic, while also retaining important information about the \textit{verb, noun}, and \textit{variable names} used in the CNL. 
Despite successfully generating NL statements and CNL propositions for 20 programs, the resulting dataset (494 pairs) is too small for effective use in data-driven approaches. To address this limitation, augmentation techniques are applied to expand the size and the diversity of the dataset.

%
%
Motivated by the existing template-based approaches~\cite{DBLP:journals/corr/abs-2010-02301,DBLP:conf/semweb/DubeyBA019,DBLP:conf/emnlp/XuRZZC018,DBLP:conf/semweb/TrivediMDL17} in creating quality datasets, we decided to apply this type of approach in our case. 
Templates ensure content consistency by generating syntactically and semantically correct data, minimizing dataset errors. They are easily modifiable and scalable, ensuring well-structured, error-free datasets with specific patterns.
%
%
%
%
Thus, we manually created CNL-NL template pairs based on the propositions of CNL2ASP, and ended up developing 369 unique templates for each grammar proposition.

\begin{table}[t]
\footnotesize
\begin{tabular}{p{0.39\linewidth}p{0.5\linewidth}}
\hline
\textbf{Sentence} & \textbf{Template} \\
\hline
CNL: \textit{Node 1} have an \textit{edge} \textit{node X}, where \textit{X} is one of \textit{2, 5.} & TCNL: \textit{Noun\_1} \textit{num\_1} have an \textit{verb\_1} \textit{ noun\_1 var\_1}, where \textit{var\_1} is one of \textit{num\_2, num\_3.}\\
NL: There is \textit{node 1} has an \textit{edge} to \textit{node X}, where \textit{X} is one of the numbers \textit{2 or 5}. & TNL: There is \textit{noun\_1 num\_1} has an \textit{verb\_1} to \textit{noun\_1 var\_1}, where \textit{var\_1} is one of the numbers \textit{num\_2 or num\_3.}\\
\hline
\end{tabular}
\caption{An overview of CNL \& NL Pair Template }
\label{tab:templatesPairs}
\end{table}

Templates were obtained as follows: we start from valid CNL-NL pair, we replace specific part of speech (nouns, verbs, etc.) with placeholders so obtain a template CNL-NL pair. 
A placeholder is a symbolic representation of a specific replacement, and we used conventions for different categories: numbers are ``num\_X", verbs are ``verb\_X", nouns are ``noun\_X", variables are ``var\_X", colors are ``color\_X", and special predicate identifiers \textit{PIDs} are ``PID\_X", and so on (``X'' is a positive integer ensuring uniqueness of placeholders).
Based on CNL patterns for specified facts, 
numeric placeholders are: (i) \textit{num\_range}, defined to signify a range of numbers; and (ii) \textit{num\_choice}, used for referencing specific numbers. 
For instance, the range from 1 to 4 is presented as \textit{num\_range (1 to 4)}, and \textit{num\_choice (1, 2, and 5)} or \textit{num\_choice (2 or 3)}.
This process is exemplified in Table~\ref{tab:templatesPairs}, where a template is obtained by introducing placeholders in sentences of a CNL-NL pair, i.e. \textit{``noun"} placeholder, used to replace the word \textit{``node"}, \textit{``verb"} placeholder replaces the word \textit{``edge"}, variable \textit{``var"} placeholder replaces capital alphabet \textit{``X"}, and number \textit{``num"} placeholders replace the numbers \textit{2 and 5.}
%
%
Numeric placeholders are accompanied by other categories like verb \textit{``verb\_X''} and noun \textit{``noun\_X''} for word variations. Placeholder \textit{``var\_X''} represents capital alphabets, while \textit{``col\_X''} denotes a list of colors. Special placeholders are introduced for contexts with the clause ``whenever,'', where the defined list of placeholders follows the pattern of \textit{PID\_X}, e.g. \textit{PID\_1} is replaced with \textit{``first vtx''}.

After creating the unique template CNL-NL pairs, we created suitable \textit{Bag-of-Words} (BoW) to obtain possible replacements for placeholders. We considered three categories of words, \textit{verbs, nouns}, and special predicate identifiers like \textit{PIDs}. We manually identified: 21 \textit{PIDs}, 77 \textit{nouns}, and 408 \textit{verbs}. Words that might be used as predicate names were taken from problems specifications, and the other were added so to construct meaningful CNL and NL.

Having defined the templates and the BoWs, we replaced the placeholders at random by respecting the placeholder expected type (e.g., \textit{verb\_X} in \textit{templates}, it is replaced by associating a verb from the BoW of \textit{verbs}), \textit{num\_X} are replaced by sequential integer values, \textit{num\_range} is replaced by numbers  generated at random within the delineated limits, \textit{num\_choice}, we choose random integers between 1 and 10 and arranged them in order and comma separated.

\begin{table}[t]
\centering
\small
\begin{tabular}{p{2.6cm}p{0.8cm}p{0.9cm}p{0.9cm}p{1.0cm}p{0.5cm}}
\hline
\textbf{Grammar Type} & \textbf{Source} & \textbf{Gener.} & \textbf{Rep.Ct.}& \textbf{Total} \\
\hline
Def. Const/Comp. & 154 & 21 & 875 & 1050 \\
Def. `When' & 145 & 15 & 800 & 960 \\
Def. `Whenever' & 110 & 41 & 755 & 906 \\
Neg. Constraint & 22 & 138 & 800 & 960 \\
Pos. Constraint & 39 & 121 & 800 & 960 \\
Quant. Choice & 13 & 156 & 845 & 1014 \\
Weak Constraint & 11 & 149 & 800 & 960 \\\hline
Grand Total & 494 & 641 & 5675 & 6810 \\
\hline
\end{tabular}
\caption{Count of CNL2ASP propositions in NL2CNL. `Gener.' stands for `Generated', `Rep.Ct.' stands for `Rephrased Count'. }
\label{tab:DatasetCount}
\end{table}

The distinct counts for each grammar proposition is reported in Table~\ref{tab:DatasetCount} in column ``Source''. To mitigate potential class imbalances in training data-driven models, we instantiated the templates in such a way that the total number of NL2CNL pairs per proposition type in the expanded dataset is almost equal. In summary, we augmented the source dataset with 641 template-generated pairs (cfr. ``Gener.'' column in Table \ref{tab:DatasetCount}), resulting in a total of 1135 NL2CNL pairs.

To enhance the quality and diversity of our dataset, we decided to perform an additional paraphrasing task on all the NL statements. 
This process increases the dataset size and introduces varied linguistic representations while maintaining the original meaning so to enable effective training of a robust Language Model (LM).
Thus, we used \textit{OpenAI API}\footnote{https://platform.openai.com/docs/api-reference} for this task, with \textit{``text-davinci-003"
} engine with specific parameters, such as: prompt set to \textit{``Rephrase the following sentence: {sentence}"}, temperature value of 0.6, and maximum token limit of 1000. Each sentence was rephrased five times in order to strike a balance between diversity and semantic dilution. Generation of more than five may lead to unintended overlaps or even slight semantic shifts. After rephrasing natural language sentences the count increased to 5675 (cfr. ``Rep.Ct.'' column in Table \ref{tab:DatasetCount}). 
Collecting together all available NL-CNL pairs we obtain the dataset NL2CNL of 6810 pairs (see Table \ref{tab:DatasetCount}).

\section{The NL2ASP tool}

\begin{figure}[t]
\centering
\includegraphics[trim={ 0 0 0 15},clip,width=0.9\columnwidth,height=3.5cm]{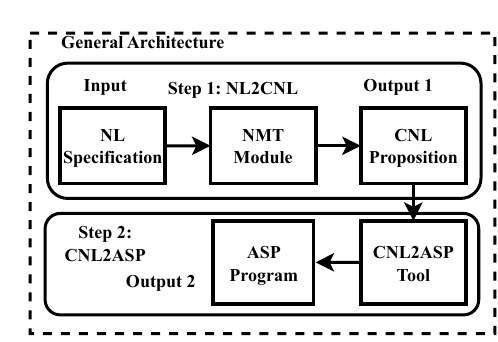} 
\caption{Architecture for automatic composition of ASP programs.}
\label{fig:NetworkOverview}
\end{figure}

The architecture of the NL2ASP tool for automatic composition of ASP programs from natural language statements is depicted in Figure~\ref{fig:NetworkOverview}. 
In the first step we transform NL statements in CNL statements, and, in the second step, we obtain the ASP code by running CNL2ASP tool.
In the proposed architecture, the first step relies on the recent advances in Neural Machine Translation~\cite{DBLP:journals/corr/abs-2002-07526}. Specifically, encoder-decoder-based Transformer architectures are incorporated to facilitate the translation of NL specifications into CNL propositions. 
Using these high-performance models, we capture language nuances and produce grammatically accurate translations. 
The use of NMT to address the first step is a decision that arises intuitively, as it is easy to realise that the task is like translating from English into another (more restricted) language, i.e., the language of CNL2ASP.
Our architecture is simple, ensuring adaptability, can be implemented with different NMT tools, and can take profit from the dataset we have presented in previous section.


In the implementation we used state-of-the-art transformer-based architectures i.e., T5~\cite{DBLP:journals/jmlr/RaffelSRLNMZLL20} and BART~\cite{DBLP:journals/corr/abs-1910-13461}, which have proven their efficiency for different language tasks like text summarization, question-answering, and machine translation.
The T5 model introduces an encoder-decoder transformer architecture that undergoes pretraining in a combination of unsupervised and supervised tasks, with each task transformed into a text-to-text format~\cite{DBLP:journals/jmlr/RaffelSRLNMZLL20}. We set T5 to implement a translation task.
BART follows a denoising autoencoder approach, where the input sequence is corrupted using a noise function~\cite{DBLP:journals/corr/abs-1910-13461}. Then, the sequence of corrupted entries passes through a bidirectional encoder, which captures contextual information from both directions. Then, the encoder representation is introduced into an autoregressive decoder from left to right, which generates the sequence by predicting the next token based on the previously generated tokens. 
\section{Experiments}
We have evaluated empirically our approach, and, in this section, describe two sub-analysis: $(i)$ assessment of the NMT model's translation quality, and $(ii)$ end-to-end assessment of the proposed architecture, i.e., assessment of the ability of our CNL2ASP to accurately create ASP programs from natural language. 
To this end, we fine-tuned pre-trained versions of the T5 and BART models on the proposed dataset and measured several evaluation metrics. 
\textit{Datasets and tools are available in the supplementary material, including additional details in appendices, that will be public upon acceptance}.

\paragraph{Software and Hardware details.}
We adopted the pre-trained T5 and BART models freely available on the Hugging Face\footnote{https://huggingface.co/} platform. 
Specifically, we used the T5-small\footnote{https://huggingface.co/t5-small} and Bart-base\footnote{https://huggingface.co/facebook/bart-base} models, which have a size of 60 and 140 million of parameters, respectively. The choice of versions was based on the capabilities of the hardware at our disposal.

The models were implemented using Keras 2.12.0 on top of Tensorflow and the Transformers 4.30.2 library~\cite{DBLP:conf/emnlp/WolfDSCDMCRLFDS20}. To run the experiments, we used an Ubuntu 20.04 server with 500GB of RAM, a 16 GB NVIDIA Tesla V100-PCIE GPU card, and CUDA 11.8. 
The code ran using Python 3.9.17 in a Jupyter Notebook environment.

\paragraph{Evaluation measures.}
The translation quality of the model was evaluated using the BLEU~\cite{DBLP:conf/acl/PapineniRWZ02}, METEOR~\cite{DBLP:conf/acl/BanerjeeL05}, and the translation \textit{Syntactic Accuracy (SA)} measures. 
BLEU is founded on precision-based attributes and functions by comparing the n-grams found in the candidate translation with those in one or more reference translations. 
BLEU falls within the range of 0 to 1, the larger the higher degree of similarity between the translation and the reference translation. The maximum order of n-grams considered is usually set to 4 (BLEU-4).
%
%
%
%
METEOR~\cite{DBLP:conf/acl/BanerjeeL05} assesses machine translation hypotheses by aligning them with one or more reference translations. This alignment process considers exact, stem, synonym, and paraphrase matches between words and phrases. The metric relies on the harmonic mean of precision and recall for unigrams. 
%
%
%
A value closer to 1 indicates a higher translation quality.
METEOR is said to have a better correlation with human judgment than BLEU.
The SA (see Equation \ref{eq:sa}) metric, measures the proportion of translated sentences that conform to the CNL2ASP grammar.
\begin{equation}\label{eq:sa}
SA= \frac{\# \text{syntantically correct sentences}}{\# \text{total sentences}}
\end{equation}

\subsection{Assessment of the NMT models}
\paragraph{Experiment\_A: Cross-validation.}
In the first experiment, we performed K-fold cross-validation, models T5 and BART are trained $``k"$ times, the dataset is divided in one fold as a test set, and $``k$-1" folds are used as the training set. This approach provides a robust estimate of model generalization ability. The value of $k$ was set to 5.
We use all 6810 NL to CNL pairs from NL2CNL dataset, which are named as features of \textit{``sentences"} and \textit{``targets"} respectively.

To perform the text tokenization, we used the \textit{AutoTokenizer} utility provided by the Transformer library, which allows us to use the pre-trained tokenizers of T5-small and Bart-base.
To handle the collating and batching of input data for training, we used the \textit{DataCollator} utility in the Transformer library. It plays a vital role in streamlining the data preparation process and ensuring seamless integration during training and inference. The batch size was set to 16.




The Adam Weight Decay (AdamW) optimizer is used in the training process.
This optimizer is based on the adaptive estimation of first and second order moments with an added method to decay the weights. AdamW is a variant of the Adam optimizer that decouples weight decay from the adaptive learning rate, which leads to better performance.
We set {learning rate=} \(2 \times 10^{-5}\), and {weight decay=} 0.01.

To calculate the BLEU we have used \textit{``corpus\_bleu"}, a function from the NLTK \cite{DBLP:journals/lre/Wagner10}
library in Python. It computes the BLEU score of the entire dataset, by taking multiple pairs of source and reference sentences. 
Moreover, we decided to calculate the cumulative BLEU up to 4-gram.
We have tuned METEOR with parameters alpha=0.9, beta=3, and gamma=0.5, which do the trade-off between recall and precision, the penalty for sentence length mismatches, and word order differences respectively.
%
We ran the training of the models for 200 epochs for each split and applied the Early Stopping technique to monitor the validation loss with minimum mode and patience equal to 20, thus reducing the overfitting and computational resources.


Table \ref{tab:bleu_meteor} illustrates the T5-small BLEU and METEOR scores by split and the final average. The BLEU scores for all n-grams are generally high, ranging from 0.90\% to 0.97\%, which shows strong performance in terms of n-gram overlap with the reference text. The METEOR score also indicates high performance, with values ranging from 0.94\% to 0.98\% across the splits.
On the other hand, Table \ref{tab:bleu_meteor_bartbase} shows the metrics scores for the fine-tuned Bart-base model. Bart-base model scores are low compared to model T5-small (cfr. Table \ref{tab:bleu_meteor}).

\begin{table}[t]
\centering
\small
\begin{tabular}{ccccccccc}
\hline
Split & BLEU-1 & BLEU-2 & BLEU-3 & BLEU-4 & MET\\
\hline

0 &	0.955 &	0.944 &	0.939 &	0.925 &	0.960 \\
1 &	0.966 &	0.957 &	0.952 &	0.940 &	0.969 \\
2 &	0.961 &	0.952 &	0.947 &	0.935 &	0.966 \\
3 &	0.959 &	0.949 &	0.944 &	0.931 &	0.961 \\
4 &	0.962 &	0.952 &	0.947 &	0.935 &	0.965 \\ \hline
Avg. & 0.961 &  0.951 &  0.946 & 0.933 & 0.964 \\

\hline
\end{tabular}
\caption{Scores across splits for T5-small model
}
\label{tab:bleu_meteor}
\end{table}

Although training times might vary with hardware, there are significant differences in training and scoring times between T5-small and Bart-base models. 
T5-small takes about 5 hours on average to train across the split, with scoring times of 3 to 5 minutes, while Bart-base averages approximately two hours for training and six hours for scoring. Despite Bart-base requiring fewer epochs, it spends more time on calculating translation quality measures.
T5-small surpasses Bart-base in all translation quality measures and excels in high-quality text generation. Despite a longer training time, T5-small delivers faster predictions compared to Bart-base. These performance advantages make T5-small the preferred choice, despite it requires longer training times.
%


\begin{table}[b]
\centering
\small
\begin{tabular}{ccccccccc}
\hline
Split & BLEU-1 & BLEU-2 & BLEU-3 & BLEU-4 & MET\\
\hline

0 &	0.841 &	0.783 &	0.751 &	0.686 &	0.875 \\
1 &	0.845 &	0.786 &	0.753 &	0.686 &	0.877 \\
2 &	0.853 &	0.797 &	0.767 &	0.704 &	0.890 \\
3 &	0.818 &	0.758 &	0.726 &	0.660 &	0.864 \\
4 &	0.842 &	0.787 &	0.757 &	0.696 &	0.879 \\ \hline
Avg. & 0.84 & 0.782 & 0.751 & 0.686 & 0.877 \\

\hline
\end{tabular}
\caption{Scores across splits for Bart-base model}
\label{tab:bleu_meteor_bartbase}
\end{table}

\paragraph{Experiment\_B: Syntax Correctness.}
This more demanding experiment aims at checking the syntactic correctness of the CNL propositions produced by the two models.
For this purpose, we measured the syntactic accuracy by subjecting each produced CNL proposition to the syntax checker provided by the CNL2ASP tool.
In particular, the syntactic correctness checking was performed during K-fold cross-validation by running the trained models on their splits. 
%
%
%
%
%
\begin{table}[t]
\centering
\small
\begin{tabular}{cccccc}
\hline
Split & Total & \multicolumn{2}{c}{Bart-base} & \multicolumn{2}{c}{T5-small} \\
\cline{3-4} \cline{5-6}
& & Errors   &  Acc. &  Errors & Acc. \\
\hline
0 &  1362 &   650 &  52.28 &   93 & 93.17 \\
1 &  1362 &   630 &  53.74 &   78 & 94.27 \\
2 &  1362 &   572 &  58.00 &   69 & 94.93 \\
3 &  1362 &   619 &  54.55 &   82 & 93.98  \\
4 &  1362 &   600 &  54.95 &   80 & 94.13 \\
\hline
\end{tabular}
\caption{
Syntax check for T5-small \& Bart-base with K=5.
}
%
\label{tab:t5bart_checksyntax}
\end{table}
Since the NL2CNL dataset has a length of 6810, for each split, a validation set of 1362 examples is used. Again model T5-small, with a SA of about 94\%, performs better than model Bart-base with SA is around 55\% (cfr. Table \ref{tab:t5bart_checksyntax}).



Looking at problematic generations we observe that T5-small exhibits issues, particularly with the clause ``whenever'' when used with other propositions. Also sentences exceeding 60 characters occasionally result in incomplete generation. Some minor issues are present in the transformation of ``less than or equal to'' to ``at most or equal to,''.
Nonetheless, a human user can fix these rare issues with minor adjustments.
On the other hand, Bart-base model exhibits some more issues in addition to those previously discussed for T5-small. These include occasional generation of missing spaces (e.g., ``C2'' generated as ``C 2''), inconsistent word connections (e.g., vice-versa), and problems with capitalization (e.g., ``whenever There is a''). Furthermore, the model occasionally uses dashes instead of underscores to connect words (e.g., ``Dominating\_Set'' vs. ``Dominating-Set''), which is not acceptable for the CNL2ASP parsing tool, but could be addressed with post-generation processing.


\paragraph{Experiment\_C: Assessment on unseen.}
In this experiment, we train both models with the complete dataset.
During training, T5 converged at 155 epochs, with a training loss of 0.022 and validation loss equal to 0.0467; the best BART model was obtained at 37 epochs, with training and validation losses of 0.010 and 0.0492, respectively.
These values are quite low, evidencing the ability of the models to learn from the dataset.
%
%
As test set we prepared a new test dataset manually, which consisted of 209 new specifications.
We did the inference and calculated BLEU,  METEOR, accuracy, precision, recall, and F1-score for both T5-small and Bart-base. T5-small consistently outperformed Bart-base in all metrics, see Table~\ref{tab:t5bart_evaluation}. 
Both models generalize well and achieve robust performance on unseen instances.

\begin{table}[b]
\centering
\small
\begin{tabular}{ccc}
\hline
Metric & T5-small & Bart-base \\
\hline
BLEU-4 & 0.860 & 0.704 \\
METEOR & 0.935 & 0.876  \\
Precision & 0.929 & 0.860 \\
Recall & 0.927 &  0.921 \\
F1-Score & 0.928 & 0.88 \\
Accuracy & 0.368 & 0.253 \\

\hline
\end{tabular}
\caption{Evaluation measures for T5-small \& Bart-base.}
\label{tab:t5bart_evaluation}
\end{table}



\subsection{End-to-End Evaluation}
We now measure the ability of the NL2ASP tool to produce valid ASP programs.
To this end, we targeted six well-known graph-related problems, i.e. Maximize Clique, Hamiltonian Cycle, Graph Coloring, Connected Dominating Set, Traveling Salesman Problem, and Hierarchical Clustering. We manually created a dataset with the specifications for the aforementioned problems, including two different encodings for the Graph Coloring problem (called GCv1 and GCv2). The inherent flexibility of natural language allows for a multitude of ways to express the same concept. So to make the assessment more realistic the dataset was expanded by paraphrasing the former specifications with the support of ChatGPT. 
%
In total, we obtained 21 problem specifications.

The NL2ASP tool, configured with the best performing t5-small fine-tuned in the Experiment\_C,  was fed with one problem specification at time to obtain the corresponding ASP programs. 
Since checking the uniform equivalence of ASP programs is intractable, we maullually-checked the correctness of the the produced programs.

During the translation step, NL2ASP was able to produce 389 syntactically correct CNL statements out of 393, performing for a good 98.98\%. In this case, we found minor errors, for example, our tool generated \textit{``at most or equal to"} instead of \textit{``less than or equal to"}. 
The mentioned issues were found in two variants of the Hierarchical Clustering problem, so NL2ASP was able to produce 19 \textit{correct} ASP programs over 21 problem specifications. 
It is important to mention that only minor manual fixes would be required by a human to fix the aforementioned issues that affected Hierarchical Clustering instances. 
(Generated CNLs statements and ASP programs can be consulted in supp. material Appendix A.).

\paragraph{Additional experience with LLMs.} As a final test, we wanted to check to what extent a general purpose LLM, such as GPT 3.5, is capable of carrying out ASP program composition. 
The goal of this experiment is not to establish a direct comparison with ChatGPT, but to ensure that specific-purpose tools like ours can perform better than general-purpose models that are not directly trained to tackle this kind of tasks.
%
The prompt we used to request the program generation is: \textit{``Please give the ASP program of the following natural language statements:”}, followed by the natural language statements.
As expected, the ``out-of-the-box'' ChatGPT was able to generate ASP programs given the specification. However, these programs are neither syntactically nor semantically correct most of the time (17/21), and quite some effort would be required to fix them manually. 
For all wrong programs we asked ChatGPT to regenerate them 3 times, but this resulted in no gain.
It is an open research question to explore how/if these models can be robustly used for this task.

\section{Conclusion}
This paper introduces the NL2CNL dataset and NL2ASP tool for automatic composition of ASP programs.
NL2ASP is based on a two-step architecture, which transforms natural language specifications in CNL statements that, eventually, are converted into valid ASP programs. 
NL2ASP has been implemented with two Transformer-based models for NMT tasks, i.e. T5-small, and Bart-base.
In our experiments, T5-small performed better than Bart-base for text-to-text translation according to several translation quality measures. 
An end-to-end analysis confirms that NL2ASP is effective and most often provides correct ASP code.
As future work, we plan to extend the dataset beyond graph-related problems, and further explore the usage of LLMs for this task.

\appendix





\bibliographystyle{named}
\bibliography{ijcai24}

\begin{thebibliography}{}

\bibitem[\protect\citeauthoryear{Alviano \bgroup \em et al.\egroup }{2017}]{DBLP:conf/lpnmr/AlvianoCDFLPRVZ17}
Mario Alviano, Francesco Calimeri, Carmine Dodaro, Davide Fusc{\`{a}}, Nicola Leone, Simona Perri, Francesco Ricca, Pierfrancesco Veltri, and Jessica Zangari.
\newblock The {ASP} system {DLV2}.
\newblock In {\em {LPNMR}}, volume 10377 of {\em LNCS}, pages 215--221. Springer, 2017.

\bibitem[\protect\citeauthoryear{Alviano \bgroup \em et al.\egroup }{2023}]{DBLP:conf/iclp/AlvianoCR23}
Mario Alviano, Davide Cirimele, and Luis Angel~Rodriguez Reiners.
\newblock Introducing {ASP} recipes and {ASP} chef.
\newblock In {\em {ICLP} Workshops}, volume 3437 of {\em {CEUR} Workshop Proceedings}. CEUR-WS.org, 2023.

\bibitem[\protect\citeauthoryear{Bahdanau \bgroup \em et al.\egroup }{2015}]{DBLP:journals/corr/BahdanauCB14}
Dzmitry Bahdanau, Kyunghyun Cho, and Yoshua Bengio.
\newblock Neural machine translation by jointly learning to align and translate.
\newblock In {\em {ICLR}}, 2015.

\bibitem[\protect\citeauthoryear{Banerjee and Lavie}{2005}]{DBLP:conf/acl/BanerjeeL05}
Satanjeev Banerjee and Alon Lavie.
\newblock {METEOR:} an automatic metric for {MT} evaluation with improved correlation with human judgments.
\newblock In {\em IEEvaluation@ACL}, pages 65--72. Association for Computational Linguistics, 2005.

\bibitem[\protect\citeauthoryear{Baral and Dzifcak}{2011}]{DBLP:conf/aaaifs/BaralD11}
Chitta Baral and Juraj Dzifcak.
\newblock Solving puzzles described in english by automated translation to answer set programming and learning how to do that translation.
\newblock In {\em {AAAI} Fall Symposium: Advances in Cognitive Systems}, volume {FS-11-01} of {\em {AAAI} Technical Report}. {AAAI}, 2011.

\bibitem[\protect\citeauthoryear{Borroto and Ricca}{2023}]{DBLP:journals/eswa/BorrotoR23}
Manuel~A. Borroto and Francesco Ricca.
\newblock Sparql-qa-v2 system for knowledge base question answering.
\newblock {\em Expert Syst. Appl.}, 229(Part {A}):120383, 2023.

\bibitem[\protect\citeauthoryear{Brewka \bgroup \em et al.\egroup }{2011}]{DBLP:journals/cacm/BrewkaET11}
Gerhard Brewka, Thomas Eiter, and Miroslaw Truszczynski.
\newblock Answer set programming at a glance.
\newblock {\em Commun. {ACM}}, 54(12):92--103, 2011.

\bibitem[\protect\citeauthoryear{Busoniu \bgroup \em et al.\egroup }{2013}]{DBLP:journals/tplp/BusoniuOPST13}
Paula{-}Andra Busoniu, Johannes Oetsch, J{\"{o}}rg P{\"{u}}hrer, Peter Skocovsky, and Hans Tompits.
\newblock Sealion: An eclipse-based {IDE} for answer-set programming with advanced debugging support.
\newblock {\em Theory Pract. Log. Program.}, 13(4-5):657--673, 2013.

\bibitem[\protect\citeauthoryear{Calimeri \bgroup \em et al.\egroup }{2020}]{DBLP:journals/tplp/CalimeriFGIKKLM20}
Francesco Calimeri, Wolfgang Faber, Martin Gebser, Giovambattista Ianni, Roland Kaminski, Thomas Krennwallner, Nicola Leone, Marco Maratea, Francesco Ricca, and Torsten Schaub.
\newblock Asp-core-2 input language format.
\newblock {\em Theory Pract. Log. Program.}, 20(2):294--309, 2020.

\bibitem[\protect\citeauthoryear{Dakhel \bgroup \em et al.\egroup }{2023}]{DBLP:journals/jss/DakhelMNKDJ23}
Arghavan~Moradi Dakhel, Vahid Majdinasab, Amin Nikanjam, Foutse Khomh, Michel~C. Desmarais, and Zhen Ming~(Jack) Jiang.
\newblock Github copilot {AI} pair programmer: Asset or liability?
\newblock {\em J. Syst. Softw.}, 203:111734, 2023.

\bibitem[\protect\citeauthoryear{Dodaro \bgroup \em et al.\egroup }{2022}]{DBLP:conf/datalog/DodaroMR22}
Carmine Dodaro, Marco Maratea, and Francesco Riccio.
\newblock A tool for encoding controlled natural language specifications as {ASP} rules.
\newblock In {\em Datalog}, volume 3203 of {\em {CEUR} Workshop Proceedings}, pages 188--201. CEUR-WS.org, 2022.

\bibitem[\protect\citeauthoryear{Dubey \bgroup \em et al.\egroup }{2019}]{DBLP:conf/semweb/DubeyBA019}
Mohnish Dubey, Debayan Banerjee, Abdelrahman Abdelkawi, and Jens Lehmann.
\newblock Lc-quad 2.0: {A} large dataset for complex question answering over wikidata and dbpedia.
\newblock In {\em {ISWC} {(2)}}, volume 11779 of {\em LNCS}, pages 69--78. Springer, 2019.

\bibitem[\protect\citeauthoryear{Eiter \bgroup \em et al.\egroup }{2007}]{DBLP:journals/tocl/EiterFW07}
Thomas Eiter, Michael Fink, and Stefan Woltran.
\newblock Semantical characterizations and complexity of equivalences in answer set programming.
\newblock {\em {ACM} Trans. Comput. Log.}, 8(3):17, 2007.

\bibitem[\protect\citeauthoryear{Erdem and Yeniterzi}{2009}]{DBLP:conf/bionlp/ErdemY09}
Esra Erdem and Reyyan Yeniterzi.
\newblock Transforming controlled natural language biomedical queries into answer set programs.
\newblock In {\em BioNLP@HLT-NAACL}, pages 117--124. Association for Computational Linguistics, 2009.

\bibitem[\protect\citeauthoryear{Erdem \bgroup \em et al.\egroup }{2016}]{DBLP:journals/aim/ErdemGL16}
Esra Erdem, Michael Gelfond, and Nicola Leone.
\newblock Applications of answer set programming.
\newblock {\em {AI} Mag.}, 37(3):53--68, 2016.

\bibitem[\protect\citeauthoryear{Ernst and Bavota}{2022}]{DBLP:journals/software/ErnstBM22}
Neil~A. Ernst and Gabriele Bavota.
\newblock Ai-driven development is here: Should you worry?
\newblock {\em {IEEE} Softw.}, 39(2):106--110, 2022.

\bibitem[\protect\citeauthoryear{et al.}{2021}]{DBLP:journals/corr/abs-2107-03374}
Mark~Chen et~al.
\newblock Evaluating large language models trained on code.
\newblock {\em CoRR}, abs/2107.03374, 2021.

\bibitem[\protect\citeauthoryear{Fang and Tompits}{2017}]{DBLP:conf/inap/FangT17}
Min Fang and Hans Tompits.
\newblock An approach for representing answer sets in natural language.
\newblock In {\em {DECLARE}}, volume 10997 of {\em LNCS}, pages 115--131. Springer, 2017.

\bibitem[\protect\citeauthoryear{Febbraro \bgroup \em et al.\egroup }{2011}]{DBLP:conf/lpnmr/FebbraroRR11}
Onofrio Febbraro, Kristian Reale, and Francesco Ricca.
\newblock {ASPIDE:} integrated development environment for answer set programming.
\newblock In {\em {LPNMR}}, volume 6645 of {\em LNCS}, pages 317--330. Springer, 2011.

\bibitem[\protect\citeauthoryear{Gebser \bgroup \em et al.\egroup }{2012}]{DBLP:series/synthesis/2012Gebser}
Martin Gebser, Roland Kaminski, Benjamin Kaufmann, and Torsten Schaub.
\newblock {\em Answer Set Solving in Practice}.
\newblock Synthesis Lectures on Artificial Intelligence and Machine Learning. Morgan {\&} Claypool Publishers, 2012.

\bibitem[\protect\citeauthoryear{Gebser \bgroup \em et al.\egroup }{2016}]{DBLP:conf/iclp/GebserKKOSW16}
Martin Gebser, Roland Kaminski, Benjamin Kaufmann, Max Ostrowski, Torsten Schaub, and Philipp Wanko.
\newblock Theory solving made easy with clingo 5.
\newblock In {\em {ICLP} (Technical Communications)}, volume~52 of {\em OASIcs}, pages 2:1--2:15. Schloss Dagstuhl - Leibniz-Zentrum f{\"{u}}r Informatik, 2016.

\bibitem[\protect\citeauthoryear{Gebser \bgroup \em et al.\egroup }{2020}]{DBLP:journals/tplp/GebserMR20}
Martin Gebser, Marco Maratea, and Francesco Ricca.
\newblock The seventh answer set programming competition: Design and results.
\newblock {\em Theory Pract. Log. Program.}, 20(2):176--204, 2020.

\bibitem[\protect\citeauthoryear{Gelfond and Lifschitz}{1988}]{DBLP:conf/iclp/GelfondL88}
Michael Gelfond and Vladimir Lifschitz.
\newblock The stable model semantics for logic programming.
\newblock In {\em {ICLP/SLP}}, pages 1070--1080. {MIT} Press, 1988.

\bibitem[\protect\citeauthoryear{Gelfond and Lifschitz}{1991}]{DBLP:journals/ngc/GelfondL91}
Michael Gelfond and Vladimir Lifschitz.
\newblock Classical negation in logic programs and disjunctive databases.
\newblock {\em New Gener. Comput.}, 9(3/4):365--386, 1991.

\bibitem[\protect\citeauthoryear{Hahn \bgroup \em et al.\egroup }{2023}]{DBLP:journals/corr/abs-2303-10118}
Susana Hahn, Orkunt Sabuncu, Torsten Schaub, and Tobias Stolzmann.
\newblock Clingraph: {A} system for asp-based visualization.
\newblock {\em CoRR}, abs/2303.10118, 2023.

\bibitem[\protect\citeauthoryear{Hua and Wang}{2020}]{DBLP:journals/corr/abs-2010-02301}
Xinyu Hua and Lu~Wang.
\newblock {PAIR:} planning and iterative refinement in pre-trained transformers for long text generation.
\newblock {\em CoRR}, abs/2010.02301, 2020.

\bibitem[\protect\citeauthoryear{Ishay \bgroup \em et al.\egroup }{2023}]{DBLP:conf/kr/IshayY023}
Adam Ishay, Zhun Yang, and Joohyung Lee.
\newblock Leveraging large language models to generate answer set programs.
\newblock In {\em {KR}}, pages 374--383, 2023.

\bibitem[\protect\citeauthoryear{Jurafsky and Martin}{2009}]{DBLP:books/lib/JurafskyM09}
Dan Jurafsky and James~H. Martin.
\newblock {\em Speech and language processing: an introduction to natural language processing, computational linguistics, and speech recognition, 2nd Edition}.
\newblock Prentice Hall series in artificial intelligence. Prentice Hall, Pearson Education International, 2009.

\bibitem[\protect\citeauthoryear{Kalliamvakou}{2022}]{github}
Eirini Kalliamvakou.
\newblock Research: quantifying github copilot’s impact on developer productivity and happiness, 2022.

\bibitem[\protect\citeauthoryear{Kuhn}{2014}]{DBLP:journals/coling/Kuhn14}
Tobias Kuhn.
\newblock A survey and classification of controlled natural languages.
\newblock {\em Comput. Linguistics}, 40(1):121--170, 2014.

\bibitem[\protect\citeauthoryear{Lewis \bgroup \em et al.\egroup }{2019}]{DBLP:journals/corr/abs-1910-13461}
Mike Lewis, Yinhan Liu, Naman Goyal, Marjan Ghazvininejad, Abdelrahman Mohamed, Omer Levy, Veselin Stoyanov, and Luke Zettlemoyer.
\newblock {BART:} denoising sequence-to-sequence pre-training for natural language generation, translation, and comprehension.
\newblock {\em CoRR}, abs/1910.13461, 2019.

\bibitem[\protect\citeauthoryear{Lierler \bgroup \em et al.\egroup }{2016}]{DBLP:journals/aim/LierlerMR16}
Yuliya Lierler, Marco Maratea, and Francesco Ricca.
\newblock Systems, engineering environments, and competitions.
\newblock {\em {AI} Mag.}, 37(3):45--52, 2016.

\bibitem[\protect\citeauthoryear{Lifschitz}{2019}]{DBLP:books/sp/Lifschitz19}
Vladimir Lifschitz.
\newblock {\em Answer Set Programming}.
\newblock Springer, 2019.

\bibitem[\protect\citeauthoryear{Lifschitz}{2022}]{DBLP:conf/iclp/Lifschitz22}
Vladimir Lifschitz.
\newblock Translating definitions into the language of logic programming: {A} case study.
\newblock In {\em {ICLP} Workshops}, volume 3193 of {\em {CEUR} Workshop Proceedings}. CEUR-WS.org, 2022.

\bibitem[\protect\citeauthoryear{Mitra and Baral}{2016}]{DBLP:conf/aaai/MitraB16}
Arindam Mitra and Chitta Baral.
\newblock Addressing a question answering challenge by combining statistical methods with inductive rule learning and reasoning.
\newblock In {\em {AAAI}}, pages 2779--2785. {AAAI} Press, 2016.

\bibitem[\protect\citeauthoryear{Moldovan \bgroup \em et al.\egroup }{2003}]{DBLP:conf/naacl/MoldovanCHM03}
Dan~I. Moldovan, Christine Clark, Sanda~M. Harabagiu, and Steven~J. Maiorano.
\newblock {COGEX:} {A} logic prover for question answering.
\newblock In {\em {HLT-NAACL}}. The Association for Computational Linguistics, 2003.

\bibitem[\protect\citeauthoryear{Nye \bgroup \em et al.\egroup }{2021}]{DBLP:conf/nips/NyeTTL21}
Maxwell~I. Nye, Michael~Henry Tessler, Joshua~B. Tenenbaum, and Brenden~M. Lake.
\newblock Improving coherence and consistency in neural sequence models with dual-system, neuro-symbolic reasoning.
\newblock In {\em NeurIPS}, pages 25192--25204, 2021.

\bibitem[\protect\citeauthoryear{Papineni \bgroup \em et al.\egroup }{2002}]{DBLP:conf/acl/PapineniRWZ02}
Kishore Papineni, Salim Roukos, Todd Ward, and Wei{-}Jing Zhu.
\newblock Bleu: a method for automatic evaluation of machine translation.
\newblock In {\em {ACL}}, pages 311--318. {ACL}, 2002.

\bibitem[\protect\citeauthoryear{Peng \bgroup \em et al.\egroup }{2023}]{DBLP:journals/corr/abs-2302-06590}
Sida Peng, Eirini Kalliamvakou, Peter Cihon, and Mert Demirer.
\newblock The impact of {AI} on developer productivity: Evidence from github copilot.
\newblock {\em CoRR}, abs/2302.06590, 2023.

\bibitem[\protect\citeauthoryear{Perevalov \bgroup \em et al.\egroup }{2022}]{DBLP:conf/semco/PerevalovDUB22}
Aleksandr Perevalov, Dennis Diefenbach, Ricardo Usbeck, and Andreas Both.
\newblock Qald-9-plus: {A} multilingual dataset for question answering over dbpedia and wikidata translated by native speakers.
\newblock In {\em {ICSC}}, pages 229--234. {IEEE}, 2022.

\bibitem[\protect\citeauthoryear{Raffel \bgroup \em et al.\egroup }{2020}]{DBLP:journals/jmlr/RaffelSRLNMZLL20}
Colin Raffel, Noam Shazeer, Adam Roberts, Katherine Lee, Sharan Narang, Michael Matena, Yanqi Zhou, Wei Li, and Peter~J. Liu.
\newblock Exploring the limits of transfer learning with a unified text-to-text transformer.
\newblock {\em J. Mach. Learn. Res.}, 21:140:1--140:67, 2020.

\bibitem[\protect\citeauthoryear{Schwitter}{2018}]{DBLP:journals/tplp/Schwitter18}
Rolf Schwitter.
\newblock Specifying and verbalising answer set programs in controlled natural language.
\newblock {\em Theory Pract. Log. Program.}, 18(3-4):691--705, 2018.

\bibitem[\protect\citeauthoryear{Stahlberg}{2020}]{DBLP:journals/jair/Stahlberg20}
Felix Stahlberg.
\newblock Neural machine translation: {A} review.
\newblock {\em J. Artif. Intell. Res.}, 69:343--418, 2020.

\bibitem[\protect\citeauthoryear{Sutskever \bgroup \em et al.\egroup }{2014}]{DBLP:conf/nips/SutskeverVL14}
Ilya Sutskever, Oriol Vinyals, and Quoc~V. Le.
\newblock Sequence to sequence learning with neural networks.
\newblock In {\em {NIPS}}, pages 3104--3112, 2014.

\bibitem[\protect\citeauthoryear{Trivedi \bgroup \em et al.\egroup }{2017}]{DBLP:conf/semweb/TrivediMDL17}
Priyansh Trivedi, Gaurav Maheshwari, Mohnish Dubey, and Jens Lehmann.
\newblock Lc-quad: {A} corpus for complex question answering over knowledge graphs.
\newblock In {\em {ISWC} {(2)}}, volume 10588 of {\em LNCS}, pages 210--218. Springer, 2017.

\bibitem[\protect\citeauthoryear{Vos \bgroup \em et al.\egroup }{2012}]{DBLP:journals/tplp/VosKOPT12}
Marina~De Vos, Doga~Gizem Kisa, Johannes Oetsch, J{\"{o}}rg P{\"{u}}hrer, and Hans Tompits.
\newblock Annotating answer-set programs in lana.
\newblock {\em Theory Pract. Log. Program.}, 12(4-5):619--637, 2012.

\bibitem[\protect\citeauthoryear{Wagner}{2010}]{DBLP:journals/lre/Wagner10}
Wiebke Wagner.
\newblock Steven bird, ewan klein and edward loper: Natural language processing with python, analyzing text with the natural language toolkit - o'reilly media, beijing, 2009, {ISBN} 978-0-596-51649-9.
\newblock {\em Lang. Resour. Evaluation}, 44(4):421--424, 2010.

\bibitem[\protect\citeauthoryear{Wolf \bgroup \em et al.\egroup }{2020}]{DBLP:conf/emnlp/WolfDSCDMCRLFDS20}
Thomas Wolf, Lysandre Debut, Victor Sanh, Julien Chaumond, Clement Delangue, Anthony Moi, Pierric Cistac, Tim Rault, R{\'{e}}mi Louf, Morgan Funtowicz, Joe Davison, Sam Shleifer, Patrick von Platen, Clara Ma, Yacine Jernite, Julien Plu, Canwen Xu, Teven~Le Scao, Sylvain Gugger, Mariama Drame, Quentin Lhoest, and Alexander~M. Rush.
\newblock Transformers: State-of-the-art natural language processing.
\newblock In {\em {EMNLP} (Demos)}, pages 38--45. Association for Computational Linguistics, 2020.

\bibitem[\protect\citeauthoryear{Xu \bgroup \em et al.\egroup }{2018}]{DBLP:conf/emnlp/XuRZZC018}
Jingjing Xu, Xuancheng Ren, Yi~Zhang, Qi~Zeng, Xiaoyan Cai, and Xu~Sun.
\newblock A skeleton-based model for promoting coherence among sentences in narrative story generation.
\newblock In {\em {EMNLP}}, pages 4306--4315. Association for Computational Linguistics, 2018.

\bibitem[\protect\citeauthoryear{Yang \bgroup \em et al.\egroup }{2020}]{DBLP:journals/corr/abs-2002-07526}
Shuoheng Yang, Yuxin Wang, and Xiaowen Chu.
\newblock A survey of deep learning techniques for neural machine translation.
\newblock {\em CoRR}, abs/2002.07526, 2020.

\bibitem[\protect\citeauthoryear{Yang \bgroup \em et al.\egroup }{2023}]{DBLP:conf/acl/YangI023}
Zhun Yang, Adam Ishay, and Joohyung Lee.
\newblock Coupling large language models with logic programming for robust and general reasoning from text.
\newblock In {\em {ACL} (Findings)}, pages 5186--5219. Association for Computational Linguistics, 2023.

\end{thebibliography}

\end{document}